\documentclass[a4paper, 10pt, conference]{ieeeconf}      

\IEEEoverridecommandlockouts                              

\usepackage[dvipdfmx]{graphicx}
\usepackage{epsfig} 
\usepackage{times} 
\usepackage{amsmath} 
\usepackage{amssymb}  

\usepackage{bm}
\usepackage{subcaption} 
\usepackage{caption}
\usepackage{multirow} 
\usepackage{subcaption} 
\usepackage{comment}
\usepackage{makecell}

\usepackage{algorithm}
\usepackage{algorithmicx}
\usepackage[noend]{algpseudocode}
\usepackage{color}
\usepackage{xcolor}
\usepackage{url}
\usepackage{caption}
\usepackage{censor} 

\newcommand{\undertitle}{This work has been submitted to the IEEE for possible publication. Copyright may be transferred without notice, after which this version may no longer be accessible}

\title{\LARGE \bf
Integration of Vision--Language--Action Control for Autonomous Forklift Unloading with Limited Real-World Data
}

\author{Koshi Oishi$^{1,*}$ and Seigo Ito$^{1}$ 
\thanks{$^{1}$ K. Oishi and S. Ito are with Toyota Central R\&D Labs., Inc. 
        \thanks{$^{*}$ Corresponding author. {\tt\small e1616@mosk.tytlabs.co.jp}}%
        }
}

\makeatletter
\newcommand{\customundertitle}{%
    \begin{center}
        \textcolor{gray}{\small \undertitle}
    \end{center}
}
\makeatother
\begin{document}

\maketitle

\customundertitle

\thispagestyle{empty}
\pagestyle{empty}
\bstctlcite{BSTcontrol}

\begin{abstract}
Forklifts play a critical role in warehousing and distribution, but existing automation systems remain limited in handling complex tasks that require sequential decision-making.
Recently, vision--language--action (VLA) models have emerged as a promising approach for multi-step decision-making by directly generating control actions from visual observations and language prompts.
Although VLA-based control has been extensively researched for manipulators, to the best of our knowledge, its application to forklifts remains unexplored.
Therefore, this study demonstrates the integration of VLA control for an autonomous forklift unloading task with limited real-world data.
Moreover, training VLA models typically requires large-scale datasets. 
As there is currently no public dataset for forklift operations, we use a VLA model pretrained primarily on robotic manipulator data and construct a training dataset via simulation, which can be scaled up at low cost.
We then adapt the pretrained VLA model to forklift control using 9.38 h of simulation data and 0.8 h of real-world demonstrations, with the latter accounting for approximately 8\% of the collected forklift data by duration.
The integrated system was validated through 72 real-world trials across four model variants.
The results show that the proposed system can accomplish an unloading task even with limited real-world data.
\end{abstract}


\section{Introduction} \label{sec:intro}
Forklifts are among the most widely used industrial vehicles in warehousing and distribution.
In the context of rapidly expanding global logistics, forklifts are indispensable material-handling equipment, and their automation has become increasingly important.
Forklift automation has already been implemented in restricted environments, using system designs broadly similar to those reported by Walter et al.~\cite{walter_Pallet_Manipulation} and Teller et al.~\cite{teller_Voice}.
However, these approaches rely heavily on accurate state estimation using multiple sensors and hard-coded logic tailored to constrained spaces ~\cite{hadwiger2026partial}.
In real-world deployments, where forklifts operate under diverse and changing conditions, hard-coding decisions for all possible situations is prohibitively costly to engineer and maintain.
This motivates the need for an alternative approach to conventional systems that depend heavily on human-engineered design.

Recently, with the rapid progress of vision--language models (VLMs), there is a growing interest in vision--language--action (VLA) models that extend VLMs to directly output actions \cite{Yin_GPT,rt1,rt2, kawaharazuka}.
VLA models address decision-making scenarios that are challenging for conventional approaches, enabling automation in long-horizon and complex tasks~\cite{kim2024openvla,openX, black2024pi_0, bjorck2025gr00t}. 
This capability can reduce reliance on hard-coded logic, which is inflexible and costly to maintain as operating conditions vary.
Applying VLA models can potentially automate decision-making in complex forklift operations.
However, most VLA research has focused on manipulators, and forklift-specific studies remain limited.

\begin{figure}[t]
	\centering
		\includegraphics[keepaspectratio, width=8.5cm, trim= 0cm 0.0cm 0cm 0.3cm,clip]{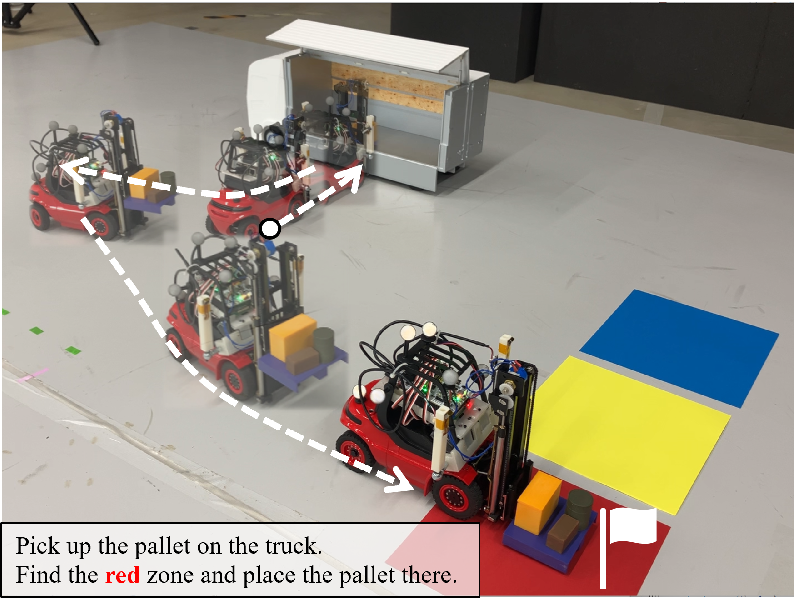}
	  \caption{Behavior of the 1/14-scale forklift executing an unloading task using our VLA model. Given only a language prompt, the forklift autonomously retrieves a pallet from the wing truck, reverses while searching the right side, and places the pallet at the location corresponding to the color specified in the prompt.}\label{fig:zu1}
\end{figure}

In this study, we integrate a VLA model into a forklift unloading system and automate the unloading sequence by directly generating actions from language and visual observations (Fig.~\ref{fig:zu1}).
Our integrated system can handle task goals in natural language and generate action sequences while jointly predicting the current task phase, eliminating the need to manually enter numeric targets or implementing extensive hard-coded logic.
Moreover, training VLA models requires large-scale datasets. 
However, to the best of our knowledge, no public dataset currently exists for forklift operations. 
Therefore, we employed a VLA model pretrained primarily on robotic manipulator data and constructed a simulation dataset so that the embodiment gap between manipulators and forklifts could be bridged without large-scale real-world forklift data.
We performed real-world evaluation of our approach on a 1/14-scale forklift platform by conducting 72 trials, including baseline comparisons.
Although full-scale forklifts are the intended target, the 1/14-scale platform enables safer, real-world iterations during early-stage learning-based development.
To the best of our knowledge, this is the first real-world integration study of VLA-based end-to-end action generation for a forklift unloading task.
The main contributions are as follows:

\begin{itemize}
    \item We integrate a VLA model into a forklift unloading system, enabling phase inference and action generation from onboard sensor observations while supporting low-latency execution.
    \item We present a cross-embodiment adaptation recipe that combines simulation-based mid-training with post-training on 0.8 h of real-world forklift demonstrations.
    \item We evaluate the proposed integrated system through 72 real-world trials across four model variants and demonstrate that it can accomplish the unloading task even with limited real-world data.
\end{itemize}

Section~\ref{sec:work} reviews related work, Sections~\ref{sec:system}--\ref{sec:data} describe the integrated system, Section~\ref{sec:demo} reports real-world experiments, and Sections~\ref{sec:lim} and \ref{sec:con} discuss limitations and conclusions.

\begin{figure*}[th]
	\centering
    \vspace*{0.2mm}
		\includegraphics[keepaspectratio, width=16cm, trim= 0cm 0.0cm 0cm 0.0cm,clip]{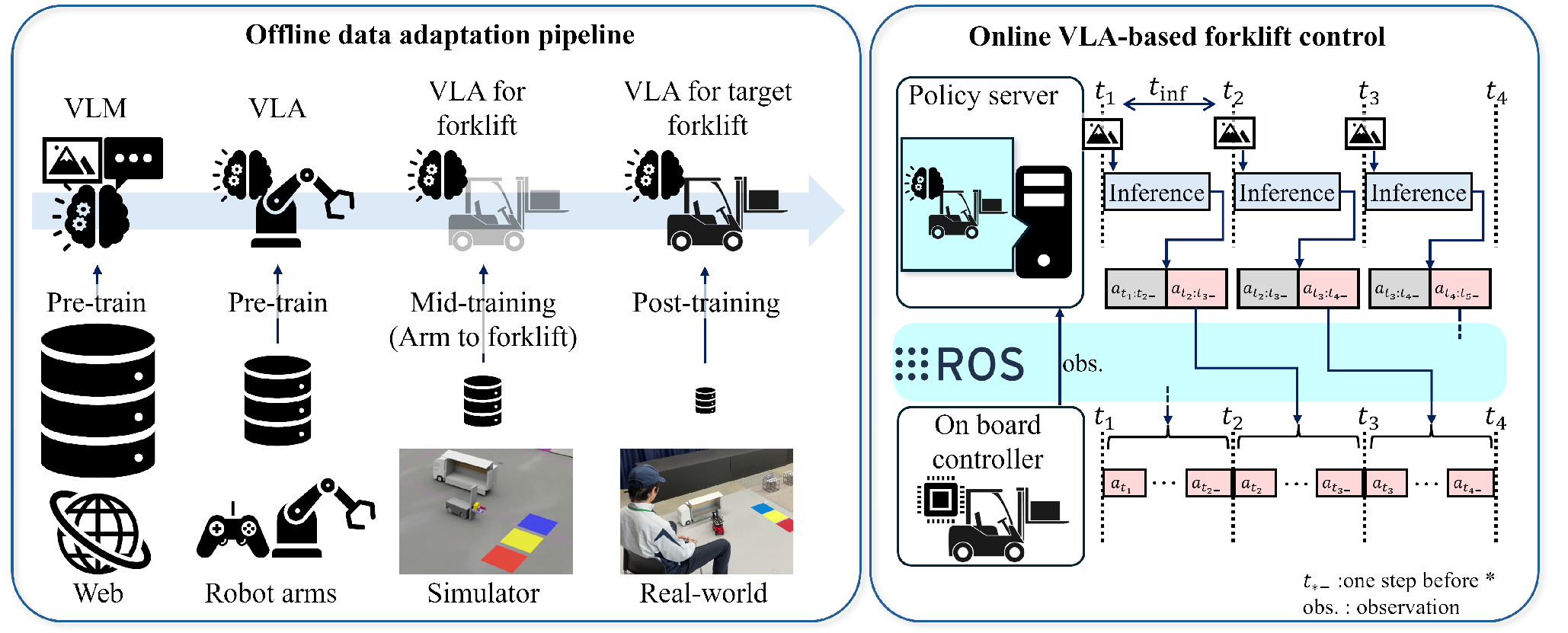}
	  \caption{Integrated system. A VLA policy is first constructed using readily available web data and manipulation data, and it is then adapted to the target forklift using simulator-generated data and a small amount of real-world data. In the online operation, inference and forklift execution run asynchronously to reduce latency. The system maintains temporal consistency by executing future action chunks predicted by the VLA.}\label{fig:system}
\end{figure*}

\begin{figure}[th]
	\centering
		\includegraphics[keepaspectratio, width=8.7cm, trim= 0cm 0.7cm 0cm 0.0cm,clip]{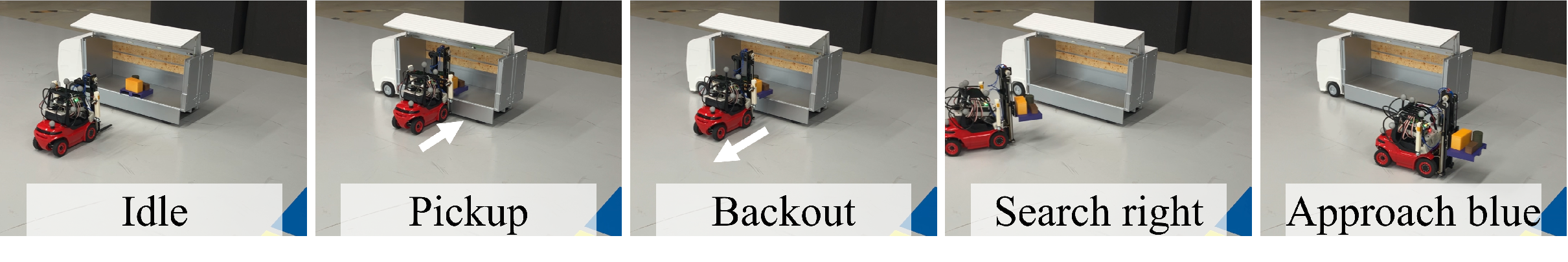}
	  \caption{Subtasks. The ``approach'' subtask shown in the last subfigure has three variants--blue, yellow, and red--giving seven subtasks in total.}\label{fig:subtask}
\end{figure}

\begin{figure}[th]
	\centering
		\includegraphics[keepaspectratio, width=6cm,trim= 0cm 0.3cm 0cm 0.0cm,clip]{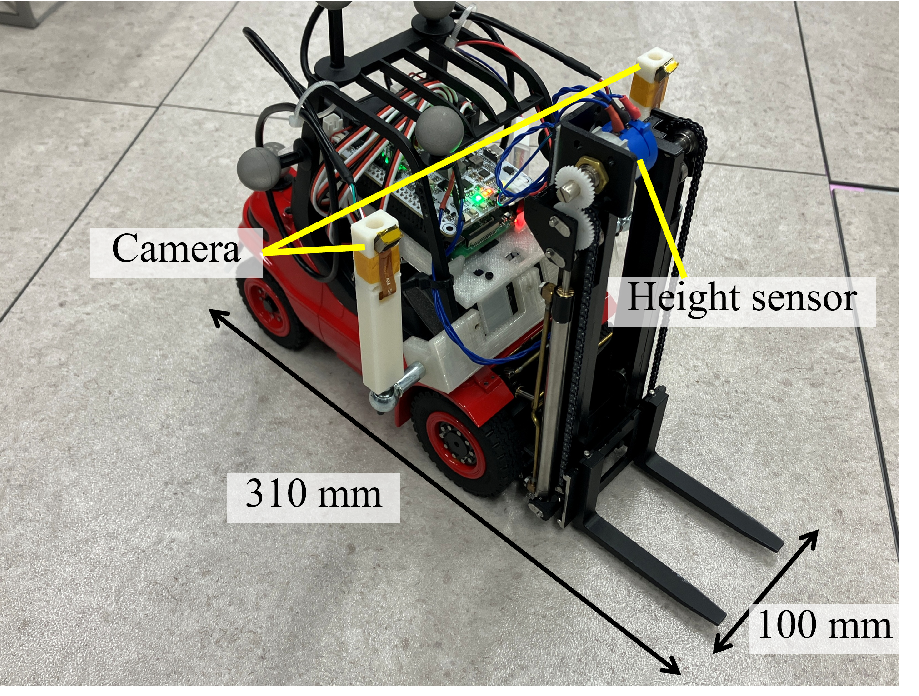}
	  \caption{1/14-scale forklift}\label{fig:hard}
\end{figure}

\section{Related work} \label{sec:work}
Recent forklift studies have combined advanced sensing technologies with deep learning to improve the robustness of environmental perception for autonomous forklift operation~\cite{Iinuma_2020jrm,huemer2025adapt, shokki_sii}. 
Nguyen et al.~\cite{nguyen2025lang2lift} used Florence-2~\cite{Xiao_florence}, a VLM, for pallet detection and target selection of forklifts.
While these studies improve perception robustness and reduce part of reliance on hard-coded logic, the transition from perception to task execution still requires manual system design.
Reinforcement-learning (RL)-based action generation has also been studied for forklift tasks~\cite{hadwiger2026partial, oishi}.
RL learns actions in a human-like manner, allowing the system to rely only on onboard sensors such as cameras and velocity.
However, RL-based studies primarily focused on short-horizon skills, such as pallet approach.
Applying RL to the entire unloading task is difficult owing to the increased complexity of reward design for long-horizon behaviors with multiple conditional branches. 
Considering these challenges, this study introduces recent VLA techniques to forklift control, thereby aiming to reduce hand-coded logic and reward design while retaining a minimal onboard sensing setup comparable to prior RL-based studies.

VLA models are commonly developed by first acquiring visual understanding and decision-making capabilities from large-scale non-robot web data, then transferring these capabilities to robot control using large-scale robot datasets, and finally fine-tuning the model on data from the target robot~\cite{kawaharazuka}.
The first two stages require large amounts of data, and collecting real-world robot data is costly and time-consuming.
Existing robot datasets are biased toward robot arms~\cite{rt1,openX} and autonomous driving~\cite{cardata}; therefore, directly applying existing VLA models to industrial machinery, such as forklifts, is challenging.
Recent advances in simulation have enabled automatic generation of high-quality robot data, and sim-to-real transfer has been demonstrated in some robotic tasks~\cite{Orbit_isaac,oishi}.
Although simulation data can be less diverse than real-world data and cannot cover all real-world conditions, it can provide useful supervision for transferring action generation to a new embodiment.
Based on this observation, we use digital-twin-generated data to adapt a VLA model pretrained on web and robot-arm data to forklift action generation, and then fine-tune it with a small amount of real-world data.
VLA models have also been developed for robot navigation~\cite{cheng2025navila}. 
Because forklift unloading requires accurate pallet alignment, pickup, and placement in addition to navigation, we use $\pi_{0.5}$~\cite{pmlr-v305-black25a}, which is pretrained on robotic manipulation data.
Additionally, simulation-based data generation automatically provides most of the subtask labels.
We demonstrate that this training recipe enables a VLA policy to perform a complex forklift unloading task even when large-scale forklift operation data are unavailable.
Moreover, the simulation environment and the 1/14-scale forklift platform are based on Oishi et al.~\cite{oishi}.

Recent VLA studies have proposed techniques for long-horizon control and real-time deployment.
For tasks consisting of multiple subtasks, methods that jointly learn task progress or semantic states with action generation have been explored~\cite{pmlr-v305-black25a,bjorck2025gr00t}. 
We integrate a simplified version of the semantic prediction approach of Black et al.~\cite{pmlr-v305-black25a} by jointly training subtask prediction with action generation. 
In addition, because VLA inference is often slower than the control cycle of industrial machinery, asynchronous action chunking has been proposed to perform inference while previously predicted actions are being executed~\cite{black2024pi_0,black2026real}. 
Our system implements a simplified version of this idea for real-time execution on a physical forklift platform.

\section{System overview} \label{sec:system}
The proposed integration combines an online VLA-based forklift control system with an offline adaptation pipeline, as shown in Fig.~\ref{fig:system}.
As a target task, we address pallet unloading from a wing truck using a 1/14-scale forklift platform for safe and repeatable evaluation.

\subsection{Task setting} \label{sec:task}
The unloading task consists of subtasks, including picking up the pallet, backing out, searching the right side for the target color while backing out, and placing the pallet at the specified color location.
Fig.~\ref{fig:subtask} illustrates the subtasks that compose the primary task.
The forklift starts approximately 0.15 m in front of the pallet.
The operator only provides the prompt; all other operations are executed autonomously.

\subsection{Hardware platform} \label{sec:hardware}
We used the 1/14-scale forklift platform introduced by Oishi et al.~\cite{oishi} (Fig.~\ref{fig:hard}).
Most hardware and sensing specifications follow those reported by Oishi et al.; therefore, we describe only the components and any deviation relevant to this study.
This platform is a counterbalance-type forklift, which is the most common forklift configuration, with rear-wheel steering and front-wheel drive.
The visual data were captured using two pinhole cameras mounted on the left and right sides of the forklift.
For the determination of the vehicle speed and yaw rate, we used a motion-capture system owing to the ease of implementation. 
In a full-scale forklift, these speeds are typically available directly from the vehicle.
Fork height was measured using a rotary encoder mounted near the top of the fork assembly, which estimated the lift height based on the amount of chain winding.
For actions, the throttle command was specified in the range 0--100\%, the steering command was defined as the servo motor angle, and the fork-height command specified the target height.
The forklift was equipped with a Raspberry Pi 4, which communicates with the inference PC via robot operating system (ROS)~\cite{ROS} to exchange action commands and sensor observations.

\subsection{Integrated architecture} \label{sec:integ}
Fig.~\ref{fig:system} shows the overall architecture of the proposed system.
The offline pipeline first trains a VLA policy on web-scale vision--language data and large-scale robot manipulation data, and then adapts it to forklift control.
Simulator-generated forklift data serves as a mid-training stage that bridges the embodiment gap between robot manipulators and the forklift, while a short post-training stage on a small amount of real-world data specializes the policy to the target platform and its operating environment.

The use of robot manipulation data is also motivated by its structural similarity to forklift operation.
Our pallet unloading task is essentially a pick-and-place problem: the forklift lifts a pallet and places it at a designated location.
For observation data acquisition, we used cameras mounted on the left and right sides of the forklift, similar to the wrist-mounted cameras on robot arms.

In online operation, the adapted policy receives camera images, vehicle-state observations, and a language prompt, and outputs an action chunk over a future horizon.
Let $t_{\rm inf}$ denote the maximum time required for one VLA inference.
Because VLA inference is computationally expensive, the system runs inference every $t_{\rm inf}$ and executes forklift actions asynchronously.
The policy is trained to predict actions up to $2t_{\rm inf}$ after the observation time, so that the controller can skip the portion corresponding to inference latency and execute the future part of the predicted action chunk.
This design reduces idle waiting time and maintains temporal consistency between observation, inference, and vehicle motion.

\begin{figure*}[t]
	\centering
    \vspace*{0.2mm}
		\includegraphics[keepaspectratio, width=14cm, trim= 0cm 0.0cm 0cm 0.3cm,clip]{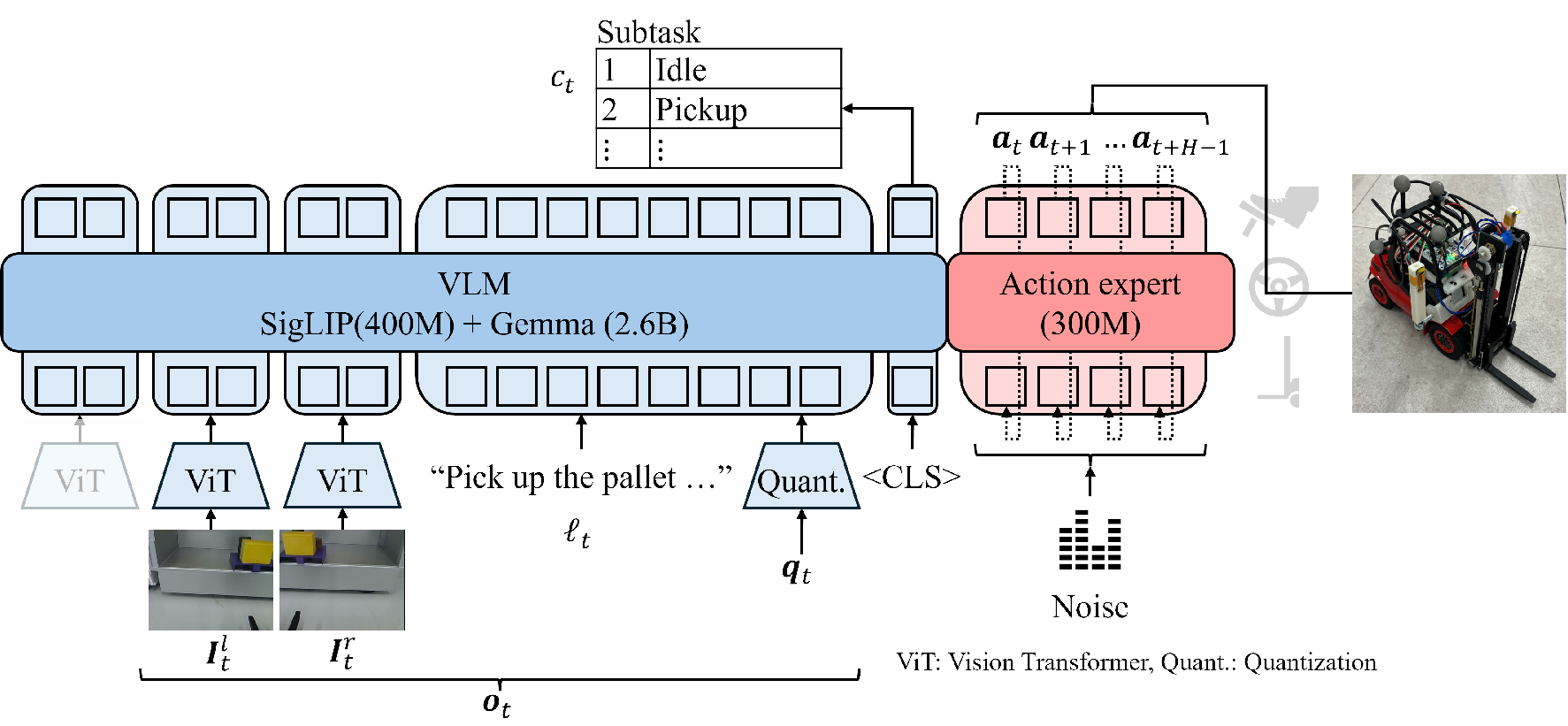}
	  \caption{VLA model reconfigured for forklift operation.}\label{fig:model}
\end{figure*}

\section{Model and training strategy} \label{sec:m_tr}
We construct the forklift policy based on the adaptation pipeline described in Section~\ref{sec:integ}.
Recent progress in robot manipulation has led to several publicly available VLA models trained on large-scale web and manipulation data.
In this work, we use $\pi_{0.5}$~\cite{pmlr-v305-black25a} as the base model, which allows us to bypass the initial construction of a VLA policy and focus on adapting it to forklift control.

\subsection{Model architecture} \label{sec:model}
$\pi_{0.5}$ is a VLA built on a PaliGemma-style VLM~\cite{beyer2024paligemma} that combines SigLIP~\cite{SigLip} and Gemma~\cite{team2024gemma}, where images are encoded by a vision transformer backbone~\cite{ViT} and actions are generated via flow matching \cite{lipman2022flow}.
An overview of our model architecture based on $\pi_{0.5}$ is shown in Fig.~\ref{fig:model}.
We denote the observation at time $t$ as $\bm{o}_t := [\bm{I}^{l}_t, \bm{I}^{r}_t, \ell_t, \bm{q}_t]$, where $\bm{I}^{l}_t$ and $\bm{I}^{r}_t$ are the left and right camera images, respectively, $\ell_t$ is the prompt, and $\bm{q}_t$ is a vector of the vehicle’s speed, yaw rate, and fork height.
These observations are compatible with onboard sensing on a full-scale forklift.
The camera images were captured at $352\times288$ RGB and resized to $224\times224$ before being fed into the VLA model.
Given the observation $\bm{o}_t$ as an input, the VLM branch classifies the current subtask, while the action-expert branch generates an action sequence $\bm{a}_{t:t+H} := (\bm a_t,\ldots,\bm a_{t+H-1})$ over a horizon $H$.
$\bm{a}_t \in \mathbb{R}^3$ includes commands throttle, steering angle, and fork height, which matches the control interface used by human operators.
The state vector $\bm{q}_t$ is quantized into 256 bins and provided to the model as text tokens.
Following the BERT design, we added a \texttt{<CLS>} token and used its feature representation for subtask classification via a multi-layer perceptron (MLP)~\cite{devlin-etal-2019-bert}.
We defined seven subtasks $c_t \in \{1,\ldots,7\}$, as shown in Fig.~\ref{fig:subtask}.
Using the same notation as $\pi_{0.5}$, our model can be described as follows:
\begin{align}
    \pi_{\theta} (\bm{a}_t, c_t | \bm{o}_t) = \pi_{\theta} (\bm{a}_t| \bm{o}_t) \pi_{\theta} (c_t| \bm{o}_t)
\end{align}
The formulation shows that the subtask prediction $c_t$ does not directly affect action generation.
Instead, the action expert attends to VLM representations that are used for subtask classification. 
During training, the classification loss backpropagates through the VLM, thereby updating the representations used by the action expert.
The base-camera input is masked because our platform lacks a head camera.

\begin{figure*}[t]
    \centering
    \vspace*{0.2mm}
        \begin{minipage}[b]{0.28\textwidth}
            \centering
            \includegraphics[width=5cm]{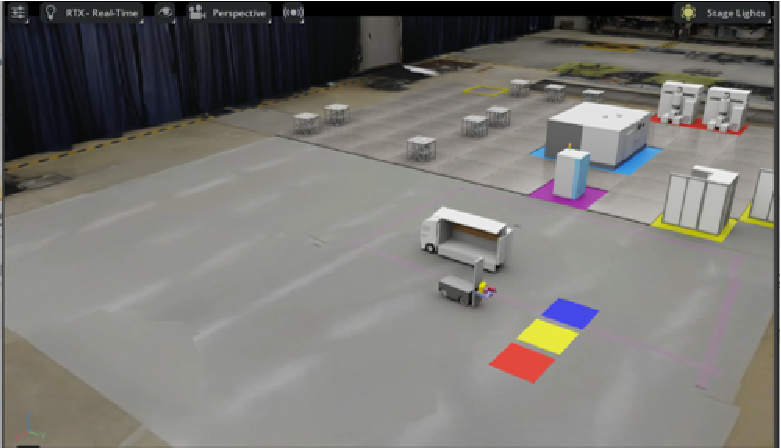}
            \subcaption{Entire environment (bright)}\label{fig:digi_env1}
        \end{minipage} 
        \begin{minipage}[b]{0.32\textwidth}
            \centering
            \includegraphics[width=5.5cm]{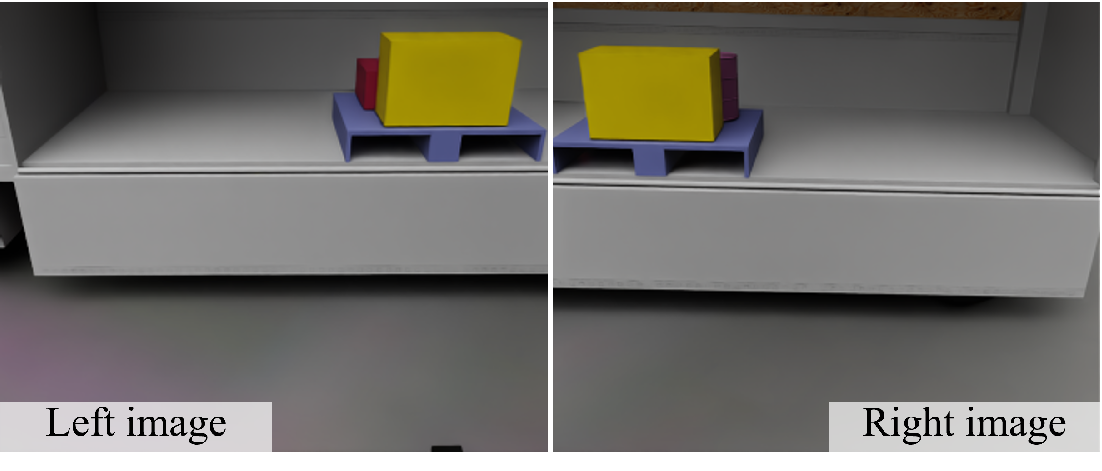}
            \subcaption{Camera image (idle)} \label{fig:cam_idle}
        \end{minipage} 
        \begin{minipage}[b]{0.32\textwidth}
            \centering
            \includegraphics[width=5.5cm, trim= 0cm 0.0cm 0cm 0cm,clip]{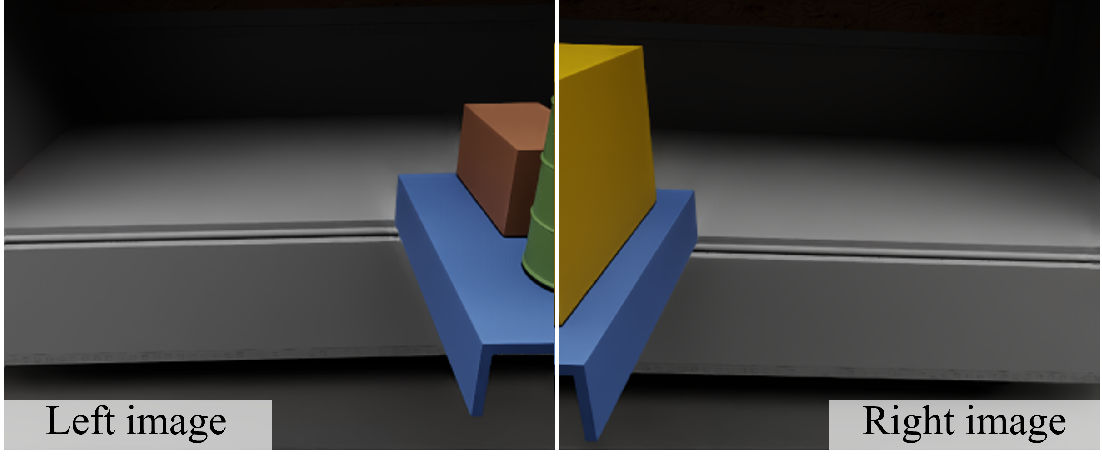}
            \subcaption{Camera image (backout)}\label{fig:cam_back}
        \end{minipage} \\
        \begin{minipage}[b]{0.28\textwidth}
            \centering
            \includegraphics[width=5cm]{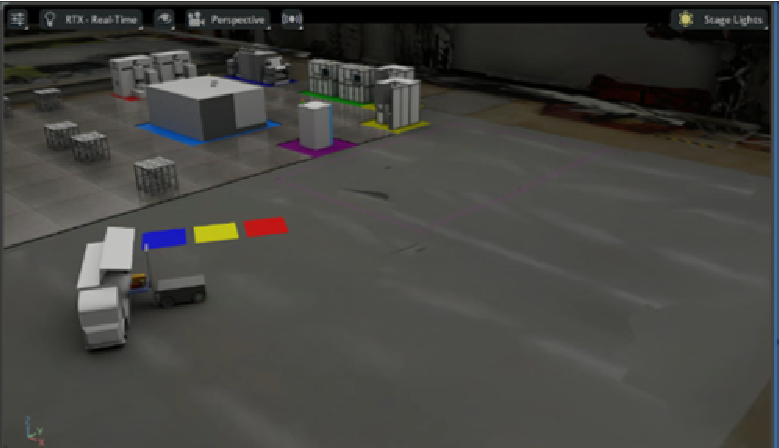}
            \subcaption{Entire environment (dark)} \label{fig:digi_env2}
        \end{minipage} 
        \begin{minipage}[b]{0.32\textwidth}
            \centering
            \includegraphics[width=5.5cm, trim= 0cm 0.0cm 0cm 0.0cm,clip]{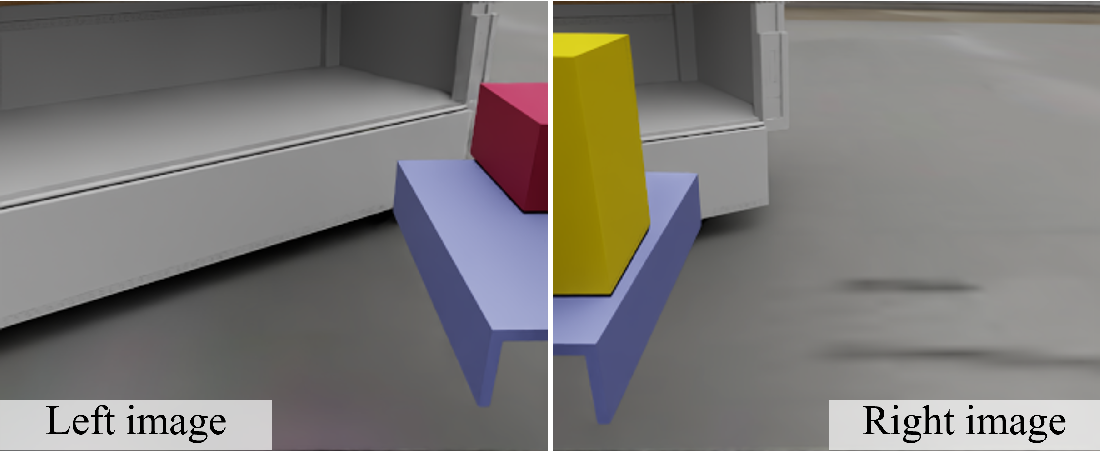}
            \subcaption{Camera image (search right)} \label{fig:cam_search}
        \end{minipage} 
        \begin{minipage}[b]{0.32\textwidth}
            \centering
            \includegraphics[width=5.5cm]{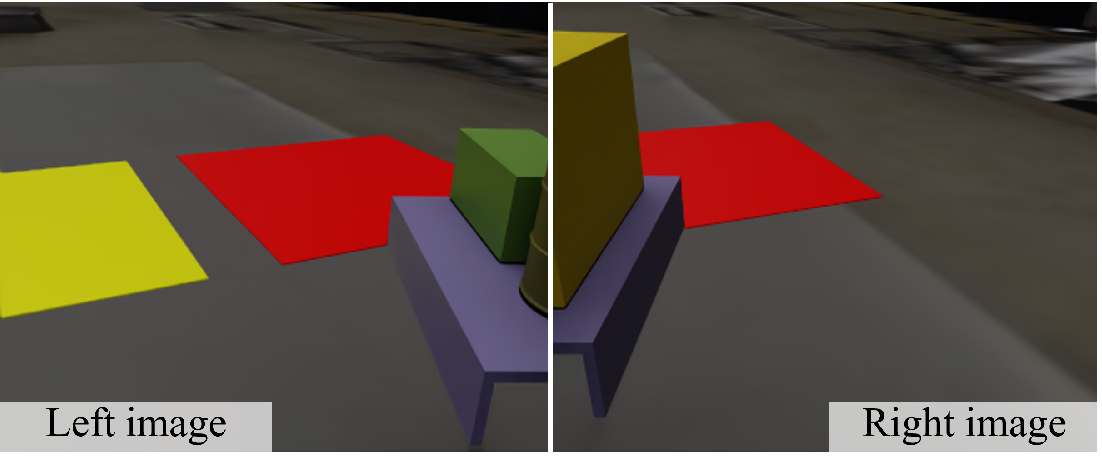}
            \subcaption{Camera image (approach red)}\label{fig:cam_app}
        \end{minipage} \\
    \caption{Environments and camera images used in simulation. (a) and (d) represent the entire environment. Position and lighting are randomized. (b), (c), (e), and (f) represent the camera images. The brackets indicate the subtask at that time.}\label{fig:sim}
\end{figure*}

\subsection{Training of action and subtask prediction} \label{sec:training}

In this study, the VLM backbone was initialized using the $\pi_{0.5}$ weights pretrained on robot data. 
The newly added \texttt{<CLS>} token and the action-expert branch were randomly initialized.
We optimized the model to minimize the following loss function:
\begin{align}
\begin{split}
\mathcal{L}(\theta)
=&
\underbrace{
\mathbb{E}_{(\bm{o}_t,\bm{a}^{*}_{t:t+H})\sim\mathcal{D}}
\left[
\ell_{\mathrm{fm}}\!\left(\theta;\bm{o}_t,\bm{a}^{*}_{t:t+H}\right)
\right]
}_{\mathcal{L}_{\mathrm{act}}}
 \\ &+
\lambda\,
\underbrace{
\mathbb{E}_{(\bm{o}_t,c_t^{*})\sim\mathcal{D}}
\left[
-\log \pi_{\theta}\!\left(c^{*}_{t}\mid \bm{o}_{t}\right)
\right]
}_{\mathcal{L}_{\mathrm{cls}}}.
\label{eq:loss}
\end{split}
\end{align}
Here, $\bm{a}^*$ and $c_t^*$ denote the ground-truth action and subtask label, respectively;  $\mathcal{D}$ denotes the training dataset; $\ell_{\mathrm{fm}}(\theta;\bm{o}_t,\bm{a}^{*}_{t:t+H})$ denotes the flow-matching loss for action sequence~\cite{black2024pi_0}; and $\lambda$ is the weight associated with the subtask classification loss.
The total loss is the sum of the action loss $\mathcal{L}_{\mathrm{act}}$ and the subtask classification loss $\mathcal{L}_{\mathrm{cls}}$.
Subtask classification provides an auxiliary semantic learning signal to the VLM representations used by the action expert. 
This objective is intended to retain task-relevant semantic information during adaptation and support consistency between the language goal and generated actions~\cite{pmlr-v305-black25a,driess2025knowledge}.

Black et al.~\cite{black2024pi_0} prepared up to 10 h of task-specific data for fine-tuning evaluation of a comparably scaled VLA model.
However, as previously indicated, forklift data are not publicly available, and collecting real-world data is expensive.
Therefore, we performed two-stage adaptation: a mid-training stage on the large-scale simulation dataset, which bridges the embodiment gap between manipulators and the forklift, followed by a short post-training stage on a small real-world dataset.
Given the expected limited diversity of the data, we applied randomization to $\bm{q}_t$ during training, as follows:
\begin{align}
    \hat{\bm{q}}_t = \bm{r}_{\mathrm{mul}} \odot \bm{q}_t + \bm{r}_{\mathrm{add}},
\end{align}
where $\hat{\bm{q}}_t$ denotes the randomized version of $\bm{q}_t$ used for training, $\bm{r}_{\mathrm{mul}}$ is the multiplicative (gain) noise, $\bm{r}_{\mathrm{add}}$ is the additive offset noise sampled from a normal distribution, and $\odot$ denotes the Hadamard (element-wise) product.
This is independent of the data-side randomization.

In this study, we used only two prompts: ``Stop and wait.'' and ``Pick up the pallet on the truck. Find the \{color\} zone and place the pallet there.''
The operator switched between these prompts and specified the target color. 
To improve robustness to observation ambiguity, we jittered the timing of prompt injection during both training and inference.
Our model uses single-shot observations without any history. 
During the forklift operation, as the camera images can remain nearly unchanged while the vehicle repeatedly moves forward and backward, selecting appropriate actions using only the current observation is challenging. 
This issue is particularly pronounced during pallet pickup, where the cameras mainly provide the inside view of the truck with little visual change.
Therefore, we split the task instruction into two sub-prompts. 
During training, we first provide ``Pick up the pallet on the truck.'' and switch to ``Find the \{color\} zone and place the pallet there.'' at the transition from the ``pickup'' subtask to the ``backout'' subtask.
During inference, this switch is triggered automatically when the fork height exceeds a specified threshold, indicating that the pallet has been lifted off the truck's floor.
This threshold-based sub-prompt switch is the only predefined logic during deployment.

\section{Dataset} \label{sec:data}
We reduced the data collection cost by generating large-scale simulation data using a simulator.
Additionally, we collected a small amount of real-world data on the target platform via human teleoperation.
Table~\ref{tb:data} summarizes the collected dataset.
In terms of total duration, real-world data accounts for approximately 8\% of the dataset.

\begin{table}[t]
\centering
\caption{Collected data}
\label{tb:data}
\begin{tabular}{ccc}
\Xhline{1pt}
Item & Simulation & Real  \\ \hline
Number of tasks executed & 1179 & 58 \\ 
Total time & 9.38 hours & 0.8 hours\\ 
Sampling period & 0.02 sec  & 0.02 sec\\ 
Number of success episodes & 898 & 58 \\ \Xhline{1pt}
\end{tabular}
\end{table}

\subsection{Simulation data} \label{sec:sim_data}
We built the simulation environment in Isaac Sim using CAD models~\cite{Orbit_isaac}. 
For the background, we used a 3D-scanned model of the actual environment.
We matched the relative configuration of the wing truck, pallet, drop-off location, and the forklift’s initial pose to the real task setup.
We then randomly placed this task set within the background model and collected $\bm{o}_t$ and $\bm{a}_t$.
To increase data diversity, we randomized the lighting intensity, the colors of the pallet and the load, the load arrangement, and the camera pose~\cite{Tobin_DR}.
Fig.~\ref{fig:sim} shows an example of the simulation environment and camera images.

During the simulation, we controlled the forklift using a hard-coded subtask manager and non-learning-based methods that leveraged privileged information available in the simulation environment (e.g., absolute pose).
We first randomly selected the target color for the drop-off location and determined the corresponding subtask sequence.
The current state of the subtask manager provided the ground-truth subtask label. 
Transitions were triggered by predefined criteria, such as fork height and reversing distance, and the subtask sequence progressed without backward transitions.
Each subtask primarily involved movement toward a designated target pose. 
To perform this movement, we used Clothoid-based path planning and Pure Pursuit for trajectory tracking \cite{paden2016survey}.
For simple behaviors such as straight ``backout'' and backing up with left steering during ``search right,'' we used a fixed steering angle.
The control period was 20 ms, and we recorded data at the same rate.
Using this setup, 9.38 h of simulation data were collected.

\subsection{Real data} \label{sec:real_data}
\begin{figure}[t]
	\centering
    \vspace*{0.2mm}
		\includegraphics[keepaspectratio, width=6cm,trim= 0cm 2.0cm 0cm 1.0cm,clip]{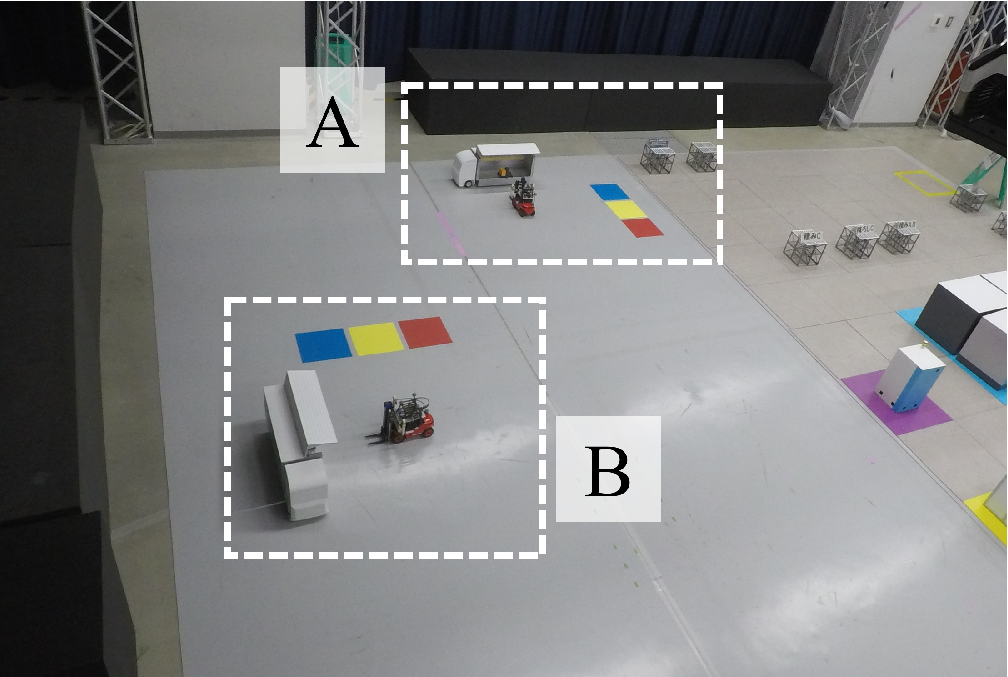}
	  \caption{Real environment. The visual background behind the pallet placement color zones differs significantly between environments A and B.}\label{fig:real_data}
\end{figure}

We collected real-world data from the two environments (A and B) shown in Fig.~\ref{fig:real_data}.
Operators teleoperated the forklift using a joystick while directly observing the scene, whereas the learned policy used onboard camera images.
Because the task does not require visual information beyond the onboard cameras’ field of view, we adopted this teleoperation method for ease of operation.
Additionally, operators manually annotated the current subtask using button inputs during teleoperation. 
The recorded button-input times defined the boundaries between successive labels.
The control period was 20 ms, which matched the simulation. All the data were logged at this rate.
In total, we collected 0.8 h of real-world data.

\section{Real-world experiment} \label{sec:demo}

We performed two-stage adaptation using large-scale simulation data, followed by a small amount of real-world data.
To assess its effect, we compared it with two baselines: a model fine-tuned based only on real-world data (``Only real'') and a model fine-tuned using only simulation data (``Only sim.'').
Table~\ref{tb:comp_model} shows the number of training steps for the simulation and real-world data for each model.
The number of real-world training updates was chosen empirically based on training progress and validation-loss trends.
We held out five real-world demonstration episodes for validation and computed the validation loss using \eqref{eq:loss} on these episodes.
In addition, we prepared a variant of our method with $\lambda = 0$ to evaluate the effect of subtask prediction. 
When $\lambda = 0$, the auxiliary subtask loss $\mathcal{L}_{\mathrm{cls}}$ in \eqref{eq:loss} was disabled, and the model was trained without subtask prediction.
Our default setting used $\lambda=1$.
Notably, the training step refers to the number of optimization updates.

\begin{table}[t]
\centering
\caption{Number of training steps}
\label{tb:comp_model}
\begin{tabular}{cccc}
\Xhline{1pt}
Item & Proposed & Only real & Only sim. \\ \hline
Simulation data & 25,000 & 0 & 25,000 \\ 
Real data & 5,000 & 10,000 & 0 \\ 
Total & 30,000 & 10,000 & 25,000 \\ \Xhline{1pt}
\end{tabular}
\end{table}

\subsection{Setting}
We conducted experiments in the two environments described in Section~\ref{sec:real_data}.
In each environment, we executed the task three times for each of the three color zones, yielding nine trials per environment.
Each model was therefore evaluated on a total of 18 trials.
Across the three trials for each color, we varied the load placement on the pallet.
A trial was considered to be successful if the forklift unloaded the pallet without dropping the load and the center of the pallet was within the target zone.
Each trial started with the prompt ``Stop and wait.''
After the operator switched the prompt to the task instruction, no further human input was provided.

\subsection{Deployment}\label{sec:dep}
As described in Section~\ref{sec:integ}, we performed inference on a ground computer.
The onboard computer applied the inferred actions to the forklift at 20 ms periods.
The ground computer was equipped with an Intel\textsuperscript{\tiny\textregistered} Xeon\textsuperscript{\tiny\textregistered} Gold 6146 processor (3.20 GHz, 24 cores), two NVIDIA RTX A6000 GPUs, and 96 GB RAM.
Even with this setup, inference took approximately 100 ms; therefore, we ran the policy at 200 ms periods ($t_{\rm inf} = 0.2$).
We trained the model to predict a sequence of 20-ms actions up to 400 ms ($H=20$).
Accordingly, we executed the predicted actions corresponding to 200--400 ms after the observation, maintaining temporal consistency between observation and control.

\subsection{Result}\label{sec:reslut}
\begin{figure}[t]
	\centering
		\includegraphics[keepaspectratio, width=8.5cm]{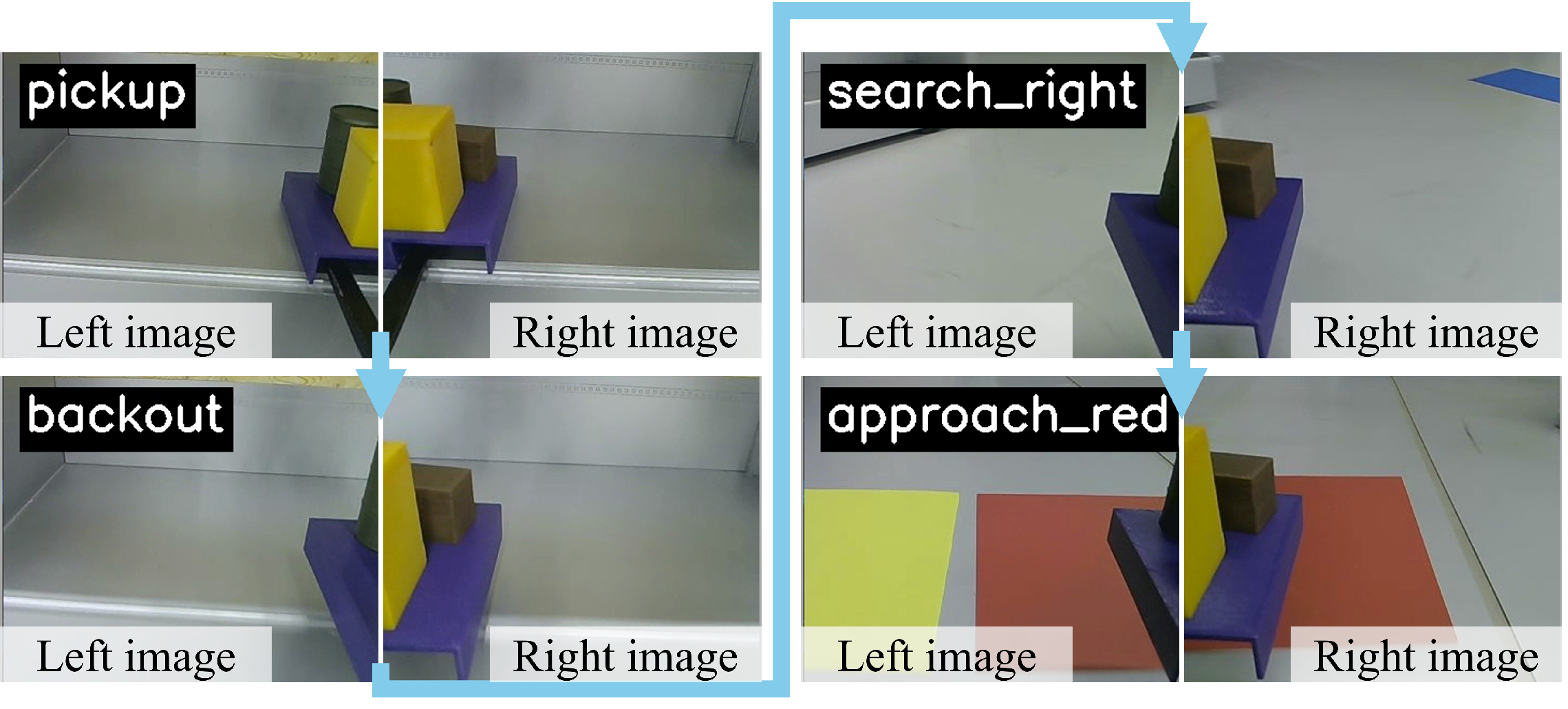}
	  \caption{Camera images in real demonstration. White texts indicate the predicted results of the current subtask.}\label{fig:real_cam}
\end{figure}

Table~\ref{tb:result} summarizes the task success rates.
The proposed method succeeded in all 18 trials. 
In contrast, ``Only real'' achieved a 50\% success rate, and ``Only sim.'' achieved 0\%.
Using the same 0.8 h real-world dataset, the proposed method achieved a higher success rate than ``Only real'' under the respective training schedules in Table~\ref{tb:comp_model}.
We evaluated subtask classification on $5$ real-world demonstration episodes comprising $12,497$ frames that were not included in the training data.
Using the operator-recorded labels, the proposed model achieved a frame-level classification accuracy of $96.8$\%.
Fig.~\ref{fig:real_cam} shows example camera images obtained as the proposed model executes the task with subtask prediction.
Fig.~\ref{fig:zu1} shows a representative successful roll-out of the proposed model.

When subtask prediction was removed ($\lambda=0$), our method achieved a lower success rate compared to ``Only real.''
The most frequent failure mode was transporting the pallet to the wrong color zone or to an unintended location.
These failures suggest that auxiliary subtask supervision helps maintain consistency between the language goal and generated actions. 
This interpretation is consistent with prior work on preserving semantic capabilities during VLA training~\cite{pmlr-v305-black25a,driess2025knowledge}.

The failure of ``Only sim.'' was likely due to insufficient data diversity, because the simulation data were generated using a single control policy.

\begin{table}[t]
\centering
\caption{Success rate}
\label{tb:result}
\begin{tabular}{cccc}
\Xhline{1pt}
\textbf{Proposed} & Proposed ($\lambda=0$) & Only real & Only sim. \\ \hline
\textbf{100 \%} & 39\% & 50 \% & 0  \% \\ \Xhline{1pt}
\end{tabular}
\end{table}

\section{Limitation}\label{sec:lim}
This study focuses on the integration of a VLA policy into a forklift unloading system under a controlled real-world setting.
Although the experiments demonstrate that the proposed system can complete the target unloading task with limited real-world data, they do not establish open-ended robustness to arbitrary environments, pallet poses, or target layouts.
Extending the evaluation to wider variations in target placement, pallet position, lighting conditions, and language instructions remains an important direction for future work.
Additionally, the high success rate highlights the potential of adapting pretrained VLA models to forklift unloading, although the specific contribution of pretraining remains to be clarified.

Although the tasks performed included a complete unloading sequence, several recovery actions required in practical logistics operations were not performed.
For example, during truck unloading, when a pallet is deep inside the truck, operators often lift the pallet while maintaining clearance from the truck, temporarily place it near the truck's edge, and then reinsert the forks more deeply to stabilize the load before extraction.
In addition, when pallets or loads interfere, fine manipulation, such as nudging the pallets with the forks to remove the interference, is often performed.
Handling such complex, contact-rich tasks that require high precision remains an important direction for future work.
For such precise operations, offline reinforcement learning is reported to be effective~\cite{intelligence2025pi}.

Finally, the experiments were conducted on a 1/14-scale forklift platform to enable safe and repeatable evaluation of the integrated system.
Deployment on full-scale forklifts will require further validation under full vehicle dynamics, stronger sensing noise, safety constraints, and collecting data.
Addressing these issues is beyond the scope of this study, however, the proposed integration provides a basis for investigating these challenges in future full-scale systems.

\section{Conclusion and future work} \label{sec:con}

This study demonstrated the integration of VLA control into a forklift unloading system.
Motivated by the challenges in automating sequential forklift operations with hard-coded logic and the lack of large-scale forklift operation data for pretraining, we adapted a pretrained VLA model to forklift control using simulation-generated and limited real-world forklift data.
The proposed integrated system connects VLA inference, phase prediction, asynchronous action-chunk execution, and low-level vehicle control.
Through 72 real-world trials, we validated that simulation-generated data support model adaptation, while subtask prediction helps align generated actions with task instructions.
The study findings demonstrate the feasibility of constructing a VLA-based forklift unloading system using simulation data, limited real-world demonstrations, and system-level integration.

\bibliographystyle{IEEEtran}
\bibliography{iros2026_3}

\end{document}